\documentclass{article}

\usepackage[preprint]{neurips_2021}

\usepackage[utf8]{inputenc} 
\usepackage[T1]{fontenc}    
\usepackage{hyperref}       
\usepackage{url}            
\usepackage{booktabs}       
\usepackage{amsfonts}       
\usepackage{nicefrac}       
\usepackage{microtype}      
\usepackage{xcolor}         
\usepackage{graphicx}       

\usepackage{hyperref}
\hypersetup{
    colorlinks=true,
    linkcolor=blue,
    urlcolor=blue,
    citecolor=black
}

\usepackage{amsthm}
\usepackage{amsmath}
\usepackage{thmtools}
\usepackage{thm-restate}
\usepackage{mathtools}

\newtheorem{definition}{Definition}

\usepackage[normalem]{ulem}

\DeclareMathOperator*{\Var}{Var}

\DeclareMathOperator*{\Lap}{Lap}

\newcommand{\cecilia}[1]{}
\newcommand{\jenny}[1]{}
\newcommand{\alex}[1]{}

\title{Combining Public and Private Data}

\author{%
  Cecilia Ferrando\textsuperscript{1,2}, Jennifer Gillenwater\textsuperscript{2}, Alex Kulesza\textsuperscript{2} \\
  \textsuperscript{1}University of Massachusetts, Amherst \\
  \textsuperscript{2}Google Research NYC \\
  \texttt{cferrando@cs.umass.com}, \texttt{\{jengi,kulesza\}@google.com} \\
}

\begin{document}

\maketitle

\begin{abstract}
   Differential privacy is widely adopted to provide provable privacy guarantees in data analysis. We consider the problem of combining public and private data (and, more generally, data with heterogeneous privacy needs) for estimating aggregate statistics. We introduce a mixed estimator of the mean optimized to minimize the variance. We argue that our mechanism is preferable to techniques that preserve the privacy of individuals by subsampling data proportionally to the privacy needs of users. Similarly, we present a mixed median estimator based on the exponential mechanism. We compare our mechanisms to the methods proposed in \cite{jorgensen2015conservative}. Our experiments provide empirical evidence that our mechanisms often outperform the baseline methods. 
\end{abstract}

\section{Introduction}
Differential privacy (DP) is a mathematical framework for drawing inferences from data while providing provable privacy guarantees for the individuals represented in the data. Informally, DP ensures that the individual contribution by a user cannot affect the aggregate statistic of interest enough to allow an adversary to infer the membership of that individual in the data. In a data-analytic context, DP is commonly achieved by adding random noise to the computation; the noise should be calibrated to the sensitivity of the output to the contribution of an individual user, and to the privacy parameter $\epsilon$, which represents how tight the privacy guarantee must be. In the literature, it is commonly assumed that $\epsilon$ is fixed for all individuals. However, in practice we can have data that comprise groups of individuals with different $\epsilon$ requirements. For example, the data might be part public, and part private, possibly with different privacy requirements. This poses the problem of how to compute a private estimator that respects all privacy requirements while also optimally capturing as much information from the data as possible.

\cite{jorgensen2015conservative} proposed ``Personalized Differential Privacy'' (PDP), as a framework for DP estimation with multiple privacy requirements. In this framework, data entries are included in the analysis according to a sampling mechanism that either includes a tuple for sure (if the corresponding $\epsilon$ is greater than a given threshold $t$) or with probability calibrated to the privacy required by the user. Their method shares similarities with \cite{alaggan2015heterogeneous}, which is instead based on rescaling data values based on the privacy requirements, but applies to a more limited set of problems (it cannot be applied to the exponential mechanism). In this paper, we propose a method for optimally mixing estimates from subgroups of the data with heterogeneous privacy requirements. In particular, we focus on the problem of computing means and quantiles from mixed datasets. Our experiments show that the proposed estimators are competitive with the PDP baseline.

\paragraph{Background}\label{background}
Formally, DP ensures that, given two ``neighboring'' data sets $X$ and $X'$ of size $n$ that differ by one entry (denoted $X \sim X'$), and given a randomized algorithm $\mathcal{A}: \mathcal{X} \rightarrow \mathbb{R}$, the probability distribution of $\mathcal{A}(X)$ is approximately the same as the distribution of $\mathcal{A}(X')$:

\begin{definition}[Differential privacy (DP), \citealt{dwork2006calibrating}]
A randomized algorithm $\mathcal{A}$ satisfies $\epsilon$-differential
privacy ($\epsilon$-DP) if, for neighboring data sets $X \sim X'$, and any subset $S \subseteq \text{Range}(\mathcal{A})$,
$$ \Pr[\mathcal{A}(X) \in S] \leq \exp(\epsilon) \Pr[\mathcal{A}(X') \in S]. $$
\end{definition}

\begin{definition}[Sensitivity, \citealt{dwork2006calibrating}]
Given any two neighboring data sets $X \sim X^{'}$, the sensitivity of a function f is $$\Delta f = \max_{X,X^{'}} \Vert  f(X) - f(X^{'}) \Vert_{1}$$
\end{definition}

The Laplace mechanism adds calibrated Laplace noise to a summary statistic of the data:
\begin{definition}[Laplace mechanism, \citealt{dwork2006calibrating}]
Given a function $f$ that maps data sets to $\mathbb{R}^{m}$, the Laplace mechanism outputs $\mathcal{L}(X) \sim \Lap(f(X), \Delta f / \epsilon)$ from the Laplace distribution, which has density $\Lap(y; u, b) = (2b)^{-m} \exp ( - \left\Vert  y - u \right\Vert_{1} / b)$. This is the same as
adding independent noise $u_i \sim \Lap(0, \Delta f / \epsilon)$ to each component of $f(X)$.  This mechanism is $\epsilon$-DP.
\end{definition}
The exponential mechanism is often used for discrete output spaces:
\begin{definition}[Exponential mechanism, \citealt{mcsherry2007mechanism}]
Given a scoring function $u: \mathcal{X} \times O \rightarrow \mathbb{R}$ with sensitivity $\Delta_u = \max_{X\sim X', o \in O} |u(X, o) - u(X', o)|$, a mechanism $\mathcal{M}(X)$ that outputs $r$ with probability proportional to $\exp(\epsilon u(X, o) / 2 \Delta_u)$ is $\epsilon$-DP.
\end{definition}

\section{Proposed mechanisms}

We consider the simple setting where all data comes from a distribution of known variance $\sigma^2$ and is defined on a range $[a, b]$.  Data is assumed to be divided into $k$ groups, with group $i$ having privacy requirement $\epsilon_i$\footnote{$\epsilon = \infty$ for public data.}.  We use $n_i$ to denote the number of data points in group $i$, a quantity that we assume is public (this is consistent with the "swap model" of DP, or the counts can be estimated privately in a separate step). Additionally, for simplicity of the exposition, we assume that each user contributes a single data point\footnote{This is not an inherent limitation of the proposed methods; they could be trivially extended with the appropriate sensitivity scaling factor to handle the case where users contribute multiple data points.}.  The problem then is how to combine data from the $k$ groups to get the best possible overall estimate of the desired statistic.  In the subsections below, we propose mechanisms for combining data to estimate two common statistics: means and quantiles.

\paragraph{Means.}

We first consider the case of computing the data's mean.  Let $\Sigma_i$ represent the sum of the data points in group $i$.  Define $r = \max(|a|, |b|)$, the max amount by which any one user can change the magnitude of their corresponding $\Sigma_i$; this is the sensitivity of the sum.  If we apply the standard Laplace mechanism to group $i$ in isolation, with $z_i \sim \Lap(r/\epsilon_i)$, then the estimate of the mean is $(\Sigma_i + z_i)/ n_i$.  What we propose is a convex combination of these individual estimates, with weights $\beta_i \geq 0$ and $\sum_{i = 1}^k \beta_i = 1$.  The joint estimator, and its corresponding variance are then:
\begin{align*}
\tilde{X} = \sum_{i=1}^{k} \beta_i \frac{\Sigma_i + z_i}{n_i}, \;\;\; s_{\text{joint}} = \sum_{i=1}^{k} \beta_i^2 \frac{n_i \sigma^2 + 2r^2/\epsilon_i^2}{n_i^2}..
\end{align*}
To find the optimal $\beta_i$, we can search for the setting that minimizes the joint variance.
This form of minimization problem has been studied before, and the Theorem in point 2 of \cite{rubin1975variance} shows that the minimizing solution is to assign each estimator a weight inversely proportional to its variance.
In more detail, for the $i$-th estimator, the optimal weight is:
\begin{align*}
    \beta_i &\propto \tilde\beta_i \coloneqq \frac{1}{\Var_i} = \frac{n_i^2}{n_i \sigma^2 + 2r^2/\epsilon_i^2}, \;\;\; \text{which normalizes to}  \;\;  \beta_i  = \frac{\tilde\beta_i}{\sum_{j=1}^{k} \tilde\beta_j} = \frac{\frac{n_i^2}{n_i \sigma^2 + 2 r^2/\epsilon_i^2}}{\sum_j \frac{n_j^2}{n_j \sigma^2 + 2 r^2/\epsilon_j^2}}\\
\end{align*}
Intuitively, the lower the variance of the estimator from one data group, the more we want that estimator to contribute to the joint estimator.  Note that in the case where one dataset is public, its corresponding $z_i$ term disappears, and the weight simplifies to $\beta_i \propto n_i / \sigma^2$.

\paragraph{Comparison with baseline method.}

In contrast to our mixing method, \cite{jorgensen2015conservative} propose a method they call the Sample mechanism. This mechanism requires first selecting a ``threshold'' hyperparameter $t$.  Then, it independently samples each data point in group $i$ with probability $\min(1, (e^{\epsilon_i} - 1) / (e^t - 1))$.  The remaining data points are fed to a DP mechanism with $\epsilon$ set to $t$, and the result of this mechanism is released.  \cite{jorgensen2015conservative} shows that this preserves the required $\epsilon_i$ privacy levels for all groups. Although this method is very flexible and can be applied to adapt any DP mechanism to the heterogeneous privacy setting, it also makes a natural baseline for our approach when applied to the Laplace mechanism. To develop some intuition for how these mechanisms compare, consider a simple example with two groups of data points, one public with $n_\mathrm{pub}$ points, and one private with $n_\mathrm{priv}$ points and parameter $\epsilon$. Under the Sample mechanism, the private points will be sampled independently as described above, whereas all public points will be retained (as they have an effective privacy parameter of $\infty$).

Suppose that a threshold $t$ is selected, and $n_\mathrm{privsamp}$ private points are subsequently sampled. Then, letting $\Sigma_\mathrm{pub}$ denote the sum of the public points and $\Sigma_\mathrm{privsamp}$ denote the sum of the sampled private points, we can write the Sample mechanism's estimate as follows:
$$
\tilde{X}_{\text{PDP}} = \alpha \cdot \frac{\Sigma_\mathrm{pub}}{n_\mathrm{pub}}  + (1-\alpha) \cdot \frac{\Sigma_\mathrm{privsamp} + \Lap(r/t)}{n_\mathrm{privsamp}} , \;\;\;\; \text{where} \;\;\alpha = \frac{n_\mathrm{pub}}{(n_\mathrm{pub} + n_\mathrm{privsamp})}.
$$
On the other hand, letting $\Sigma_\mathrm{priv}$ denote the sum of \emph{all} private points, our method computes
$$\tilde{X} = \beta \cdot \frac{\Sigma_\mathrm{pub}}{n_\mathrm{pub}} + (1-\beta) \cdot \frac{\Sigma_\mathrm{priv} + \Lap(r/\epsilon)}{n_\mathrm{priv}}
$$
for the value of $\beta$ that minimizes the overall variance. Note that, for $\tilde{X}_\mathrm{PDP}$, the choice of threshold $t$ determines both the sampling rate for the private points {\it and} the coefficient $\alpha$. In particular, if $t = \epsilon$, then all private points are sampled and the second terms in the two estimates above coincide. However, $\alpha$ will not be equal to $\beta$---which is the variance-minimizing choice---and therefore the variance of the Sample mechanism will be suboptimal.

\alex{please check the last bit for clarity...i think it's correct but it got a little hairy. if anyone has ideas for simplifying it, go for it.}

On the other hand, if $t \neq \epsilon$, then the variance of the second term of $\tilde{X}_\mathrm{PDP}$ will itself generally be suboptimal. To see this, recall that the probability with which each private point is sampled is $\min(1, (e^{\epsilon} - 1) / (e^t - 1))$, which is at most $\hat{p} \coloneqq \min(1, \epsilon / t)$ since $(e^x - 1)/x$ is increasing and therefore $(e^{\epsilon} - 1) / (e^t - 1) \leq \epsilon / t$ whenever $\epsilon \leq t$. In expectation, then, at most $n_\mathrm{priv}\hat{p}$ private points will be sampled. In that case, the variance of the second term will be at least
$$
\frac{n_\mathrm{priv}\hat{p}\sigma^2 + 2 (r/t)^2}{(n_\mathrm{priv}\hat{p})^2}
=
\frac{\sigma^2}{(n_\mathrm{priv}\hat{p})} + 2 \left(\frac{r/t}{n_\mathrm{priv}\hat{p}}\right)^2
\geq 
\frac{\sigma^2}{n_\mathrm{priv}} + 2 \left(\frac{r/\epsilon}{n_\mathrm{priv}}\right)^2~,
$$
where for the first term we use the fact that $\hat{p} \leq 1$, and for the second we use the fact that $\hat{p} \leq \epsilon/t$. Note that the right hand side is now exactly the variance of the Laplace mechanism applied to the private points alone, i.e., the second term of $\tilde{X}$. Thus, when $t \neq \epsilon$ the Sample mechanism is a convex combination of two independent terms, where the first term matches our mechanism but the second term has higher variance. Since our mechanism selects the variance-minimizing parameter $\beta$, there can be no value of $\alpha$ that compensates for this deficit.

\paragraph{Quantiles.} Private quantiles are typically computed via the exponential mechanism (see Algorithm 2 in \cite{smith2011stoc}).  \cite{jorgensen2015conservative} proposes a PDP instance of the exponential mechanism (the $\mathcal{PE}$ mechanism) that combines all data groups and their various $\epsilon_i$ into one utility function.  We propose instead to simply run an independent exponential mechanism for each data group, using the standard utility function from \cite{smith2011stoc}.  We then re-use the mixing weights derived for the computation of means in order to mix the results from these exponential mechanisms to produce a single overall quantile estimate.  We compare our mixing method against the $\mathcal{PE}$ mechanism in a simple setting (Section~\ref{Experiments}). The experiments show that our proposed mixing strategy is competitive with the $\mathcal{PE}$ mechanism in RMSE performance. We leave it to future work to develop a mixing strategy tailored explicitly for the exponential mechanism (as opposed to re-using the mixing weights that are optimized for mean estimation).

\section{Experiments}\label{Experiments}
\begin{figure}[h]
    \centering
    \includegraphics[width=\linewidth]{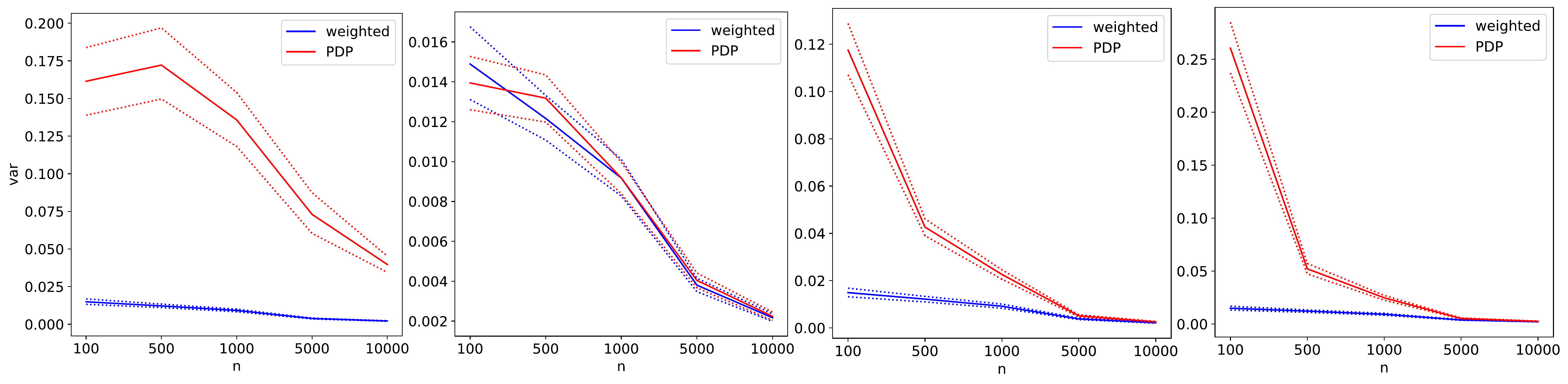}
    \caption{Variance of the weighted vs PDP mean estimators over 1000 trials. Dotted lines represent 95\% CIs. $\vec{n} = [n, 100, 500, 1000, 5000, 10000]$ and $\vec{\epsilon}=[10.0, 0.05, 0.1, 0.01, 0.25, 0.15]$. Left to right, the PDP threshold $t$ is set to: $t_{\text{min}} (0.01)$; $t_{\text{optimized}} (0.25)$; $t_{\text{average}} (1.76)$; $t_{\text{max}} (10.0)$. $t_{\text{opimized}}$ is the optimal threshold obtained by minimizing the PDP joint variance. In this case, the performance of PDP matches that of our proposed mechanism. In the other cases, the variance of the proposed estimator is always lower than the variance of the PDP estimator, with the two converging as $n \rightarrow \infty$.}
    \label{fig:multi}
\end{figure}

\paragraph{Means.} We compare our proposed estimator to the PDP Sample method from \cite{jorgensen2015conservative} in a scenario with $k$ data subgroups --- one very low-privacy group with $\epsilon_{\text{max}} = 10$ and $n$ data points, and the remaining higher-privacy groups with varying privacy requirements and sizes. Data is drawn from a normal distribution with $\mu = 0$ and $\sigma^2 = 25$. We fix the size of the private groups and let $n$ vary with values from $100$ to $10000$. For the threshold value $t$, we follow \cite{jorgensen2015conservative} and compare setting $t$ equal to the minimum $\epsilon$ among the groups, or the average of the $\epsilon$ values across groups. We also directly optimize $t$ for minimum variance, which leads to a similar performance as our weighted estimator. Additionally, we test the case where $t = \epsilon_{\text{max}}$. Results in Figure~\ref{fig:multi} show the competitive advantage of our proposed estimator over PDP, with a lower overall variance.

\begin{figure}[h]
    \centering
    \includegraphics[scale=0.3]{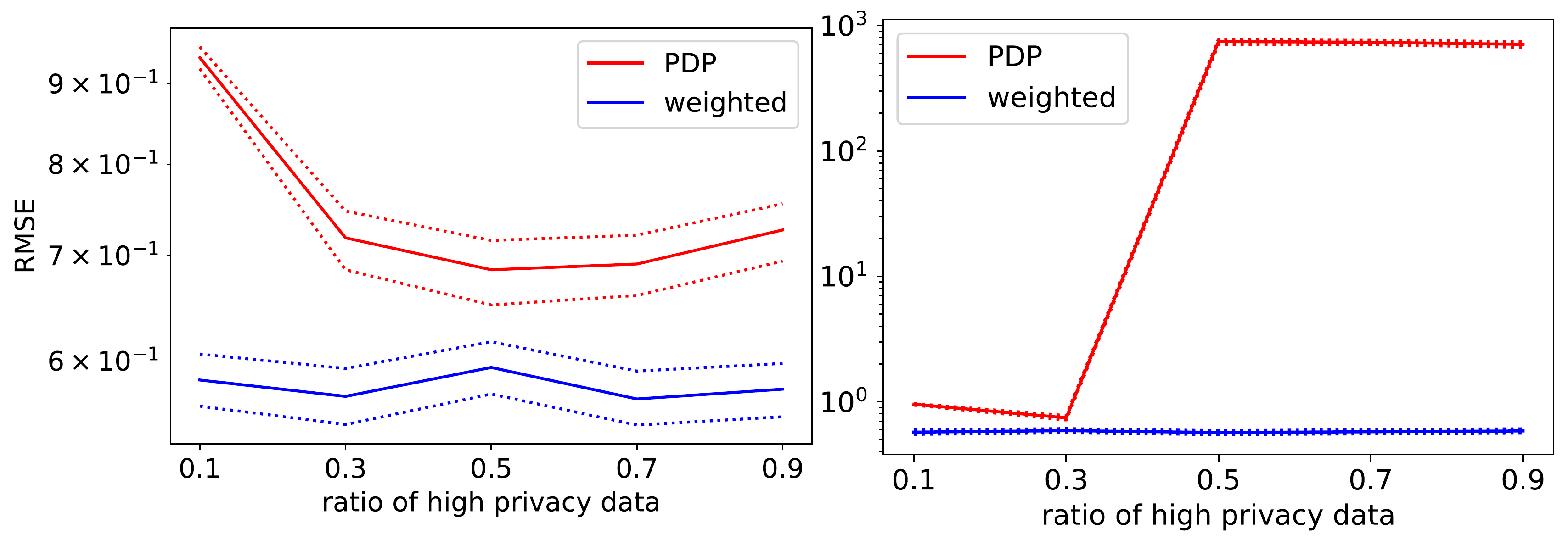}
    \caption{RMSE for the median of a standard normal over 500 trials. Left: $\epsilon_H = 0.1$, $\epsilon_L = 1.0$. Right: $\epsilon_H = 0.01$, $\epsilon_L = 10.0$.}
    \label{fig:medians}
\end{figure}

\paragraph{Medians.}We also consider the release of medians. In particular, we consider the simple setting where part of the data comes from users who require a high level of privacy, $\epsilon_H = 0.1$, and the rest from users with a looser privacy constraint of $\epsilon_L = 1.0$. In a second scenario, we look at $\epsilon_H = 0.01$ and $\epsilon_L = 10.0$. Data from both groups is drawn from a standard normal distribution with total (odd) number of data points, $n$, set to $1001$  \jenny{This is kind of a strange number.  Why not just $1000$?}\cecilia{for this experiment I followed the same setup as in the PDP paper (1001 was to have an odd number of points). I'm rerunning it on a standard normal as well.}\jenny{Why is an odd number of points needed?}and we compare different ratios of high vs low-privacy users. We compare the RMSE (root mean squared error) of the median released via the $\mathcal{PE}$ mechanism, and our weighted median. Results in Figure~\ref{fig:medians} show the competitive advantage of the weighted median. 

\section{Discussion and conclusions}
We present a minimum-variance unbiased estimator of the mean in the case of heterogeneous data with multiple privacy requirements. Our method is based on a weighting scheme that can also be applied for the release of quantiles. Our estimators often outperform the existing baseline of Personalized Differential Privacy~\citep{jorgensen2015conservative}. One limitation of our mechanism is that it adds random noise to every subgroup. In the worst case where every one of the $k$ groups has a single user and the $\epsilon_i$ differ only by an infinitesimal amount, we end up adding $k$ times as much noise as PDP Sample. Future work will explore this tradeoff, as well as a custom weighting scheme optimized for the exponential mechanism.

\newpage
\bibliography{main}

\begin{thebibliography}{6}
\providecommand{\natexlab}[1]{#1}
\providecommand{\url}[1]{\texttt{#1}}
\expandafter\ifx\csname urlstyle\endcsname\relax
  \providecommand{\doi}[1]{doi: #1}\else
  \providecommand{\doi}{doi: \begingroup \urlstyle{rm}\Url}\fi

\bibitem[Jorgensen et~al.(2015)Jorgensen, Yu, and
  Cormode]{jorgensen2015conservative}
Zach Jorgensen, Ting Yu, and Graham Cormode.
\newblock
  \href{http://dimacs.rutgers.edu/~graham/pubs/papers/pdp.pdf}{Conservative or
  liberal? Personalized differential privacy}.
\newblock In \emph{International Conference on Data Engineering (ICDE)}, 2015.

\bibitem[Alaggan et~al.(2015)Alaggan, Gambs, and
  Kermarrec]{alaggan2015heterogeneous}
Mohammad Alaggan, S{\'e}bastien Gambs, and Anne-Marie Kermarrec.
\newblock \href{https://arxiv.org/pdf/1504.06998.pdf}{Heterogeneous
  differential privacy}.
\newblock \emph{arXiv preprint arXiv:1504.06998}, 2015.

\bibitem[Dwork et~al.(2006)Dwork, McSherry, Nissim, and
  Smith]{dwork2006calibrating}
Cynthia Dwork, Frank McSherry, Kobbi Nissim, and Adam Smith.
\newblock
  \href{https://link.springer.com/content/pdf/10.1007/11681878_14.pdf}{Calibrating
  noise to sensitivity in private data analysis}.
\newblock In \emph{Conference on Theory of Cryptography (TCC)}, 2006.

\bibitem[McSherry and Talwar(2007)]{mcsherry2007mechanism}
Frank McSherry and Kunal Talwar.
\newblock \href{http://kunaltalwar.org/papers/expmech.pdf}{Mechanism design via
  differential privacy}.
\newblock In \emph{Foundations of Computer Science (FOCS)}, 2007.

\bibitem[Rubin and Weisberg(1975)]{rubin1975variance}
Donald~B Rubin and Sanford Weisberg.
\newblock
  \href{https://onlinelibrary.wiley.com/doi/pdf/10.1002/j.2333-8504.1974.tb00860.x}{The
  variance of a linear combination of independent estimators using estimated
  weights}.
\newblock \emph{Biometrika}, 62\penalty0 (3):\penalty0 708--709, 1975.

\bibitem[Smith(2011)]{smith2011stoc}
Adam Smith.
\newblock
  \href{http://cs-people.bu.edu/ads22/pubs/2011/stoc194-smith.pdf}{Privacy-preserving
  statistical estimation with optimal convergence rates}.
\newblock In \emph{Symposium on the Theory of Computing (STOC)}, 2011.

\end{thebibliography}
\end{document}